\def\year{2020}\relax
\documentclass[letterpaper]{article} 
\usepackage{aaai20}  
\usepackage{times}  
\usepackage{helvet} 
\usepackage{courier}  
\usepackage[hyphens]{url}  
\usepackage{graphicx} 
\usepackage{graphicx}  
\title{Modeling Psychotherapy Dialogues with Kernelized Hashcode Representations:\\
A Nonparametric Information-Theoretic Approach
}
\author{
Sahil Garg$^1$,
Irina Rish$^2$,
Guillermo Cecchi$^2$,
\\
sahil.garg.cs@gmail.com,
\{rish, gcecchi\}@us.ibm.com,
\\
\\
{
\normalsize
\textbf{Palash Goyal}$^1$,
\textbf{Sarik Ghazarian}$^1$,
\textbf{Shuyang Gao}$^1$,
\textbf{Greg Ver Steeg}$^1$ \and
\textbf{Aram Galstyan}$^1$
}
\\
{
\normalsize
\{palashgo, sarikgha, gaos\}@usc.edu,
\{galstyan, gregv\}@isi.edu
}
\\
\\
$^1$ USC Information Sciences Institute
\\
$^2$ IBM T. J. Watson Research Center
\\
}

\newcommand{\shrink}[1]{}

\usepackage{times}
\usepackage{soul}
\usepackage[utf8]{inputenc}
\usepackage{graphicx}
\usepackage{amsmath}
\usepackage{booktabs}
\usepackage{algorithm}
\usepackage{algorithmic}

\usepackage{xcolor}

\usepackage{url}
\usepackage{verbatim}
\usepackage{amssymb}
\usepackage{amsthm}
\usepackage{subfigure}
\usepackage{epstopdf}
\usepackage{lipsum}
\usepackage{color}

\definecolor{shadecolor}{gray}{0.95}
\newcommand{\algshade}[1]{
\hspace*{-\fboxsep}
\colorbox{shadecolor}{
\parbox{\linewidth}{
#1
}}
}

\newcommand{\csize}{
\fontsize{8}{8}\selectfont
}

\newcommand{\csizenine}{
\fontsize{9}{9}\selectfont
}

\newcommand{\todo}[1]{\textcolor{blue}{\bf{TODO}:#1}}

\newcommand{\cC}{{\mathcal{C}}}

\newcommand{\cI}{{\mathcal{I}}}

\newcommand{\cH}{{\mathcal{H}}}

\newcommand{\cT}{{\mathcal{T}}}

\newcommand{\bs}[1]{\boldsymbol{#1}}

\newcommand{\thmref}[1]{Theorem~\ref{#1}}
\newcommand{\tabref}[1]{Tab.~\ref{#1}}
\newcommand{\figref}[1]{Fig.~\ref{#1}}

\newcommand{\secref}[1]{Sec.~\ref{#1}}

\newcommand{\algoref}[1]{Alg.~\ref{#1}}

\newtheorem{theorem}{Theorem}

\begin{document}

\maketitle

\begin{abstract}
We propose a novel dialogue modeling framework, the first-ever \emph{nonparametric kernel functions} based approach for dialogue modeling, which learns {\em kernelized hashcodes} as compressed text representations; unlike traditional deep learning models, it handles well relatively  {\em small datasets}, while also scaling to large ones. We also derive a novel {\em lower bound on mutual information}, used as a model-selection criterion favoring representations with better alignment between the utterances of participants in a {\em collaborative dialogue} setting, as well as higher predictability of the generated responses.
As demonstrated on three real-life datasets, including prominently psychotherapy sessions, the proposed approach significantly outperforms several state-of-art neural network based dialogue systems, both in terms of computational efficiency, reducing  training time   from days or weeks to hours, and the response quality, 
achieving an  order of magnitude improvement over competitors in frequency of being chosen as the best model by human evaluators.
\end{abstract}

\section{Introduction}
Dialogue modeling and generation is an active research area  of great practical importance  as it provides a solid basis for building successful conversational agents in a wide range of applications.  However, despite recent successes of deep neural dialogue models,   the open dialog generation problem is far from being solved~\cite{vinyals2015neural,li2016deep,serban2016building,serban2017hierarchical,li2017adversarial,zhao2018unsupervised,wu2018learning,zhang2018generating,xu2018diversity,park2018hierarchical,jiang2018sequence,pandey2018exemplar,tao2018get,xing2018hierarchical,chen2018hierarchical,shen2018improving,shi2018toward,du2018variational,li2019insufficient,gupta2019insights}.

Therefore, it is important to continue exploring novel types of models   and  model-selection criteria, beyond today's deep neural dialogue systems, in order to better capture the  structure of different types of dialogues and to overcome certain limitations of neural models, including dependence  on large training datasets, long training times, and difficulties incorporating non-standard objective functions, among others. Along those lines, in this paper, we propose a \emph{first-ever nonparametric approach}~\cite{wasserman2006all}, i.e. based on \emph{convolution kernel similarity} functions~\cite{haussler1999convolution,mooney2005subsequence,scholkopf2001learning}, for modeling and generation of dialogues.
                    
Moreover, different applications may possess specific properties which suit some approaches better than others.
%
%
%
In this work, one of the motivating applications is a fast-growing area of  (semi-)automated \emph{psychotherapy}:  easily accessible, round-the-clock psychotherapeutic services provided by a conversational agent. The importance of this area cannot be underestimated:   according to recent statistics, mental health disorders affect one in four adult Americans, one in six adolescents, and one in eight children; predicted by the World Health Organization, by 2030 the amount of worldwide disability and life loss attributable to depression may become greater than for any other condition, including cancer, stroke, heart disease, accidents, and war. 

However, many people do not receive an adequate treatment. One of the major factors here is  limited availability of mental health care professionals, as compared to the number of potential patients; thus, automating at least some aspects of the treatment is a promising  direction.
%
%


One of the domain-specific challenges in automated therapy is {\em difficulty obtaining large training datasets} which are often necessary for neural dialogue models; this limitation may require developing alternative approaches.   Another domain-specific property  of therapeutic dialogues, which can  potentially  simplify dialogue generation, is the classical pattern of relatively long patient's utterances (up to thousands of words) followed by much shorter  therapist's responses.   Therapist's responses are often high-level, generic statements,   confirming and/or summarizing  patient's responses; they {\em can be viewed as semantic ``labels" to be  predicted from patient's ``input samples"}.
    


\begin{figure}[tp!]
\centering
\includegraphics[
width=\columnwidth]{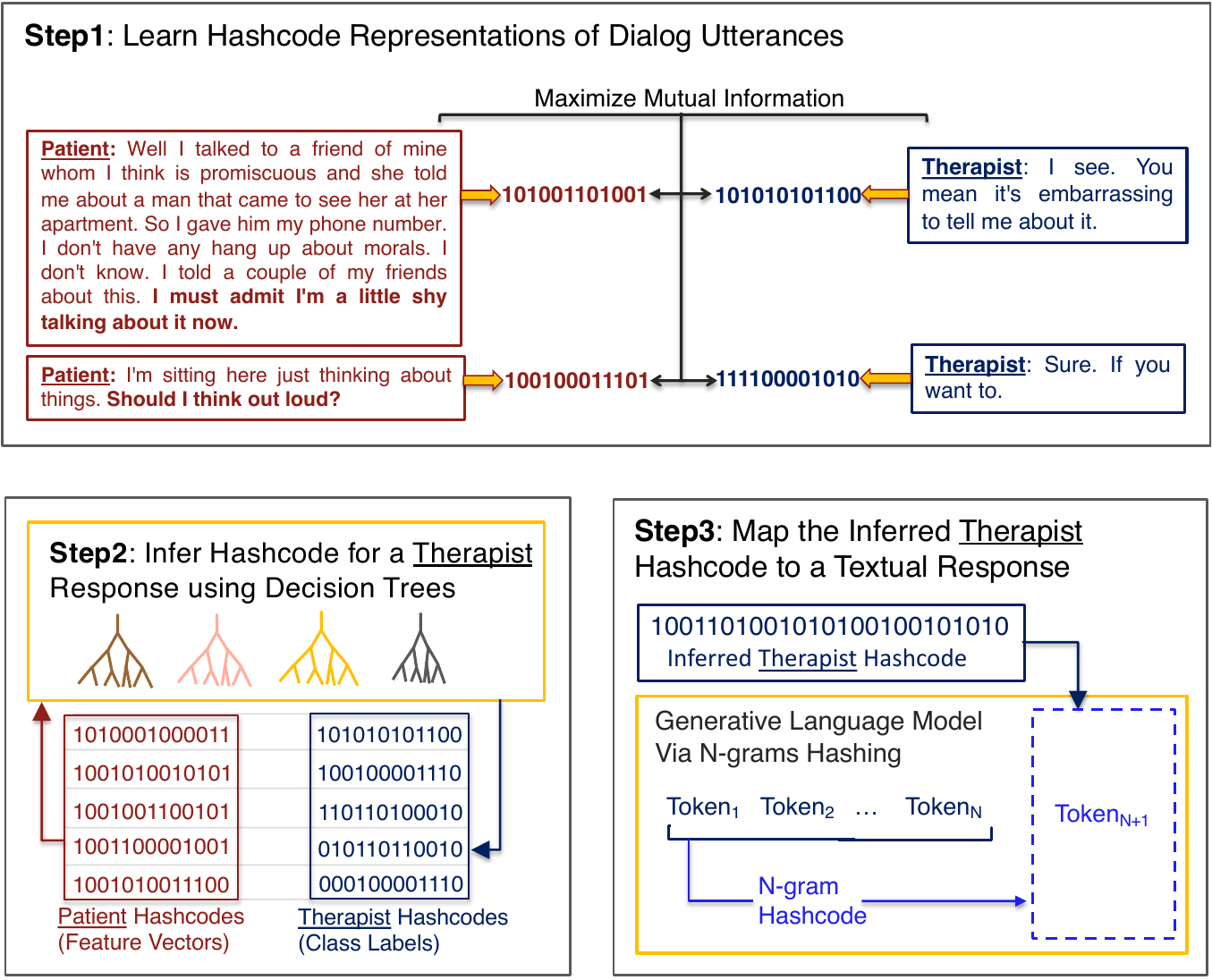}
\label{fig:hash_app}
\vspace{-5.25mm}
\caption{
\csize
An intuitive illustration of the kernelized hashcodes based framework for dialog modeling. It involves three steps. In the first step~(top), we learn a hashing model to obtain hashcode representations of dialogue utterances; the objective behind learning a hashing representation via maximizing the  mutual information between a patient (left) and therapist (right) responses is to find a compressed encoding of those responses which preserves the mutually relevant content while ignoring irrelevant details; e.g., in the example above, we would expect a good representation model to capture  the content highlighted in  boldface as essential to the conversation.
In step 2~(bottom left), we train a ensemble of decision tree classifiers so as to infer hashcode for a therapist response, given the hashcode representation of the corresponding patient utterance~(input).
In step 3~(bottom right), having the inferred therapist hashcode, we map it to a textual response. One can choose a response from the training set of therapist responses, the one for which its hashcode has low hamming distance w.r.t. the inferred therapist hashcode. Or, we can generate a novel response using N-Grams based language modeling, generalized with kernelized hashing of N-Grams.
%
}
\vspace{-4.75mm}
\label{fig:klsh_max_mi}
\end{figure}

Furthermore, a therapy session is  typically  an example of a   {\em collaborative dialogue}, unlike  debates, political arguments, and so on.
Indeed, a fundamental concept in psychotherapy is the working alliance between the therapist and the patient~\cite{bordin1979generalizability}. The alliance involves 
 the agreement on the goals to be achieved and the tasks to be carried out, and the bond, trust and respect to be established over the course of the therapy. While an encompassing formalization of working alliance is a challenging task, we propose {\em maximizing  mutual information between the patient's and therapist's responses} as a simple criterion helping us capture, to some extent, the dynamics of agreement expected to develop in most therapies, and, more generally, in other types of dialogues\footnote{Note that imbalanced response length between the two participants, as well as collaborative property, are   shared with some other types of dialogues, e.g., TV show interviews such as Larry King dataset analyzed in this paper, where the guest of a show produces long responses, with the host inserting relatively short comments facilitating the interview.}. Furthermore, {\em  maximizing mutual information}  between the patient's and therapist's can  {\em improve the   predictability of the latter from the former},  thus facilitating better dialogue generation.
                
Motivated by above considerations, we introduce here  a novel dialogue modeling framework, the \emph{first-ever nonparametric approach for dialogue modeling}, where responses are represented as \emph{locality-sensitive kernelized hashcodes}~\cite{kulis2009kernelized,joly2011random,garg2018efficient,garg2019nearly}, and the hashing models are optimized using a {\em novel mutual-information lower bound}, since exact mutual information computation is intractable in high-dimensional spaces; one can also use neural networks for learning the hashcode representations, as we demonstrate in the experiments section. Using hashcode representations allows for a more tractable way of  predicting responses in a compressed, general representation space instead of direct generation of textual responses. (Previously, kernelized hashcode representations were  successfully applied in the prior works on  information-extraction   \cite{garg2018efficient,garg2019nearly}.) Once the compressed representation of the response is inferred, any separately trained generative model can be plugged in to produce an actual textual response. See \figref{fig:klsh_max_mi} for an illustration of the framework.
%
%
Note that separating response inference in the representation space from the actual text generation increases method's flexibility, while mutual information criterion facilitates better alignment between the responses of two subjects and higher predictability of the proper response. 
It is also important to note that,  while the  psychotherapy domain was our primary motivation, the proposed approach is generally   applicable to a wider range of   domains as demonstrated in empirical section. 

Overall, our  \emph{key contributions} include:
(1) a novel  generic framework for dialogue modeling and generation using \emph{kernelized locality sensitive hash functions};
(2) a \emph{novel lower bound on the Mutual Information~(MI)} between the hashcodes of the responses from the two agents used as an optimization criterion for the locality sensitive hash functions; (3) a language model to generate therapist responses, generalizing from the traditional N-grams based language modeling approach via kernelized hashing of N-grams.
                                
We provide an extensive empirical evaluation on three different dialogue domains, from depression therapy to TV show interviews and Twitter data,  demonstrating {\em advantages   of our approach} when compared with the state-of-art neural network based dialog systems, both in terms of the {\em higher quality of generated responses} (especially on relatively small datasets (thousands of samples), including therapy sessions and Larry King TV interviews), as well as   {\em computational efficiency, reducing the training time from days or even weeks (e.g. on near-million-sample Twitter dataset) to a few hours}.

\section{
Dialog Modeling via Kernelized Hashcode Representations
}
We now present a novel framework for  dialogue modeling using binary hash functions. We will refer to the two dialogue agents as to a patient and a therapist, respectively, although the approach is generally applicable to a wider  variety of dialogue settings, as demonstrated later in the empirical section on datasets such as TV show interviews and Twitter dialogues.

\subsection{Problem Formulation and Approach Overview}
We consider a dialogue dataset consisting of $N$ samples, $\bs{S}^{pt} = \{ S_i^p, S_i^t \}_{i=1}^N$, where each sample is a pair of a patient  and a therapist responses,  $S_i^p$ and $S_i^t$, respectively; we will also use the following notation: $\bs{\bar{S}}= \{ S_1^p, \cdots, S_N^p, S_1^t, \cdots, S_N^t \}$, $\bs{S}^p= \{ S_1^p, \cdots, S_N^p \}$, $\bs{S}^t= \{ S_1^t, \cdots, S_N^t \}$. Each response is a natural language structure which can be simply a text, or a text   with part of speech tags (PoS), or a syntactic/semantic parsing of the text.

Given response  $S_i^p$, the dialogue generation task is  to  produce the  response $S_i^t$. We approach this task as a three-stage problem: first, we learn a representation model, based on {\em locality sensitive hashing}, which maps each text response $S_i$ into some binary hashcode  vector   $\bs{c}_i  \in \{0, 1\}^{H}$; second, we train a classifier to infer   the therapist's hashcode $\bs{c}_i^{t*}$ given the patient's hashcode $\bs{c}_i^p$, so that the inference takes place in the abstract representation space; \emph{hashcode representation  aims   at capturing, in a compressed form, the semantic essence of the responses while leaving out irrelevant details}; finally, we produce a   textual response based on the predicted hashcode representation.
%
    
Our objective is to choose  a hashcode-based text representation model 
so that consecutive responses of the dialogue participants are maximally relevant to each other,  as measured by the  mutual information between the corresponding hashcode representations; from another perspective, this will also make the response of the  second's person   more predictable given the first person's response. 

\subsection{Background: Kernelized Locality Sensitive Hashing}
\label{sec:lsh}
The main idea behind locality sensitive hashing is that, data points which are similar to each other as per some features, are assigned hashcodes within a short Hamming distance to each other, and vice versa  \cite{grauman2013learning,zhao2014locality,wang2017survey}. Such hashcodes can be used as  generalized representations of language structures, e.g. responses of dialogue participants.
There are multiple  hash functions   proven to be locality sensitive \cite{wang2017survey}. Kernelized locality-sensitive hashing approaches have also been developed~\cite{kulis2009kernelized,joly2011random}, which are recently shown to be applicable for learning representations of natural language \cite{garg2018efficient,garg2019nearly}. These techniques rely  on a convolution kernel similarity  function $K(S_i,S_j; \bs{\theta})$ defined for any pair of structures $S_i$ and $S_j$ with kernel parameters $\bs{\theta}$ \cite{srivastava2013walk,mooney2005subsequence,haussler1999convolution}. 
%
%
%
%


In order to construct hash functions for  mapping textual responses to   hashcodes,   we will first select from a training dataset a random subset of text structures (responses) $\bs{S}^R \subset \bs{\bar{S}}$ of size $|\bs{S}^R|=M$, called a {\em reference set}. Further, let $h_l(S_i)$, $l=1,\cdots, H$, denote a set of $H$ binary-valued hash functions, and  
let $\bs{h}(S_i)$ denote  vector $\{h_l(S_i)\}_{l=1}^H$.  The hashcode representation of response $S_i$ will be given as $\bs{c}_i=\bs{h}(S_i)$. 
    
To generate a hash function $h_l(S_i)$, for each bit $l$, we first select a random  subset  $\bs{S}^R_l \subset \bs{S}^{R}$ of the reference set,  $|\bs{S}^R_l|=2\alpha$. Next, we assign label 0 to $\alpha$ randomly selected elements of $\bs{S}^R_l$, and label 1 to the remaining $\alpha$ elements of that set, creating an artificial binary-labeled training dataset~\footnote{For a more sophisticated methodology to assign the artificial binary labels, and to select the random subsets of $\bs{S}^{R}$, refer to the recent work by \cite{garg2019nearly}.}, which can be now fed into any binary classifier to learn a function $h_l(S_i)$. 
%
%
We generate $H$ such random splits of the reference set, and learn the corresponding $H$ binary classifiers, as hash functions. We also tried  kernelized k-nearest neighbor classifier (kNN),  resulting into hashing approach  we refer to as {\em LSH-RkNN}.

Overall, to obtain a hashcode of a given response $S_i$,  we must compute $M$  kernel similarities, $K(S_i, S_j),  \forall S_j \in \bs{S}
^R$. For a limited  size~($M$) of the reference set $\bs{S}^R$,  hashcodes can be computed  efficiently, with the computational cost linear in $H$; also, note that LSH techniques described above are easily  parallelizable.
                
Finally, our {\em LSH-RLSTM} model uses LSTM language model   for generating hashcodes;  no reference set optimization is required here, since  LSTM  easily handles large training datasets; however, other hyperparameters, including    network's architecture,  need to be optimized.



\subsection{Learning Kernelized Hashcode Representations for Dialog Modeling}
\label{sec:oa}

Given that each specific hashing model described above  involves several model-selection choices, our task  will be to optimize those choices using the  information-theoretic criterion proposed below.

\subsubsection{Optimizing LSH Model Parameters\\}
As per the discussion of LSH above, an LSH model involves the function $\bs{lsh}(.; \bs{\theta}, \bs{S}^R)$ for mapping text responses to hashcodes: %
$
\bs{c}_i = 
\bs{lsh}(S_i; \bs{\theta}, \bs{S}^R)$, where 
$\bs{c}_i = \bs{h}(S_i) = (h_l(S_i))_{l=1}^H,$
and where each hash function $h_l(.)$ is built based on a random subset of $\bs{S}^R$ using either a kernel (kNN, SVM) or a neural network (LSTM) classifier. For the case of kernel-based LSH,  $\bs{\theta}$ are the parameters of a convolution kernel similarity function $K(S_i, S_j)$. For neural hashing (LSH-RLSTM), $\bs{\theta}$ refers to the  neural architecture hyperparameters (number of layers, the number of units in a layer, type of units, etc.); $\bs{\theta}$ also includes LSH-specific parameters such as  $\alpha$. 

When learning LSH models on a training datset, the (hyper) parameters   $\bs{\theta}$ as well as the reference set $\bs{S}^R$  will be optimized with respect to the information-theoretic objective introduced below.
Namely, for LSH-RkNN and LSH-RMM, the kernel parameters $\bs{\theta}$ are optimized via grid search. For LSH-RLSTM, $\bs{\theta}$ reflects the neural architecture, i.e. the number of layers and the number of units in each  layer, optimized by greedy search. Similarly, $\bs{S}^R$ is also constructed via a greedy algorithm. 
%
%

\begin{algorithm*}[tp!]
\caption{Response Generation via Hashing of N-grams}
\begin{algorithmic}[1]
\REQUIRE Therapist response sentences from the training set, $\bs{S}^t$ and their hashcodes, $\bs{C^t}$; N-grams from the training set, $\bs{S^g}$, and their hashcodes, $\bs{C^g}$; an inferred therapist hashcode, $\bs{c^t_*}$.
\STATE $S^t \gets$ samplePrefix($\bs{c^t_*}, \bs{C^t, \bs{S}^t}$)
\COMMENT{sampling from therapist responses matching the inferred therapist hashcode.}
\WHILE{True}
\STATE $S^g \gets$ extractSuffixAsNgram($S^t$)
\\
\algshade{
\STATE $\bs{c}^g \gets$ computeHashcodeOfNgram($S^g$)
%
\STATE $w^g \gets$ inferNextToken($\bs{c}^g, \bs{C^g}$)
\COMMENT{for an Ngram, there are tokens to follow as suffix along with the probabilities. Tokens from all the relevant N-Grams in the training set, matched via hashcodes, are aggregated for sampling.}
}
\STATE $S^t \gets$ appendTokenToResponse($S^t, w^g$)
\\
\algshade{
\STATE $\bs{c}^t \gets$ computeHashcodeOfResponse($S^t$)
%
\STATE $f^t \gets$ computeLikelihood($\bs{c}^t, \bs{c}^t_*, \bs{C^t}$)
\COMMENT{likelihood per the responses in training set matched via hashcodes, and  hamming distance w.r.t. the inferred hashcode.}
}
\IF{$f^t$ is high}
\RETURN $S^t, f^t$
\COMMENT{return the therapist response.}
\ELSIF{$f^g$ is very low}
\algshade{
\STATE $S^t \gets$ backtrack($S^t$)
\COMMENT{Some of the tokens in the suffix are removed from the response.}
}
\ENDIF
\ENDWHILE
\end{algorithmic}
\label{alg:response_generation}
\end{algorithm*}

\subsubsection{Information-Theoretic Objective Function\\}
The objective function for hashcode-based model selection in dialog generation should (1) characterize the quality of hashcodes as generalized/compressed representations of dialogue responses and  (2) favor  representation models leading to higher-accuracy response generation.
    
Mutual information $\cI(S^p : S^t)$ between the dialog responses $S^p$ (e.g, patient) and $S^t$ (e.g., therapist) is a natural candidate objective as it implies   higher predictability of one response from another. Though,  it   is hard to compute in practice as the joint distribution over all pairs of textual responses is not available. However, we can attempt to approximate it using hashcode representations. If $\bs{h}(.)$ represents a function from the space of all statements to the hashing code space, then the {\em data processing inequality} implies that $\forall \bs{h}(.), \cI(S^p : S^t) \geq \cI(\bs{h}(S^p) : \bs{h}(S^t))$, and   maximizing the quantity on the right can be more computationally feasible.
    
Thus we  will maximize the \emph{mutual information (MI)} between the response hashcodes, over LSH model parameters; it turns out that MI  reflects   both the inference accuracy as well as the representation quality, as we will see below:

\begin{align}
&
\mathop{\arg\max}_{\bs{\theta}, \bs{S}^R}\ \ 
\cI(\bs{C^p}:\bs{\cC^t});
\\
&
\bs{C^p}=\bs{lsh}(S^p; \bs{\theta}, \bs{S}^R),\ 
\bs{C^t}=\bs{lsh}(S^t; \bs{\theta}, \bs{S}^R)
\\
&
\cI(\bs{C^p}\!:\bs{C^t}) 
=
\cH(\bs{C^t})
- 
\cH(\bs{C^t}|\bs{C^p})
\label{eqn:dialogues_mi}
\end{align}
	
\noindent Herein, $\bs{C^p}$ and $\bs{C^t}$ are the multivariate binary random variables associated with the hashcodes of patient and therapist   responses, respectively. Minimizing the conditional entropy, $\cH(\bs{C}^t|\bs{C}^p)$,   improves the predictive accuracy when inferring  therapist response hashcode, while maximizing the entropy term, $\cH(\bs{\cC}^t)$, should ensure good quality of the hashcodes as generalized representations of text responses; thus MI objective satisfies both criteria stated at the beginning of this section.

\begin{table*}[tp!]
\centering
%
%
\subtable[Depression Therapy Dataset]
{
\centering
\renewcommand{\arraystretch}{1.4}
\renewcommand{\tabcolsep}{12.75pt}
\begin{tabular}{llllllllllll}
\toprule	
%
\textbf{Model}
& Average
& Greedy
& Extrema
\\
\toprule
%
%
%
LSTM~\protect\cite{vinyals2015neural}
& 0.61$\pm$0.31
&\bf{0.58$\pm$0.29}
&0.28$\pm$0.16
\\
\midrule
HRED\protect\cite{serban2016building}
&0.48$\pm$0.23
&0.43$\pm$0.20
&0.29$\pm$0.16
\\
\midrule
VHRED\protect\cite{serban2017hierarchical}
&0.48$\pm$0.23
&0.43$\pm$0.20
&0.29$\pm$0.16
\\
\toprule
LSH-RkNN
%
%
%
&0.60$\pm$0.37
&\bf{0.58$\pm$0.34}
&\bf{0.37$\pm$0.24}
%
%
%
%
%
%
%
\\
\midrule
LSH-RMM
%
&0.56$\pm$0.38
&0.53$\pm$0.33
&0.31$\pm$0.23
\\
%
%
%
%
%
%
\midrule
LSH-RLSTM
&\bf{0.64$\pm$0.37}
&0.51$\pm$0.28
&0.28$\pm$0.19
\\
\toprule
\end{tabular}
}
%
%
\subtable[
Twitter Dataset
]
{
\centering
\renewcommand{\arraystretch}{1.4}
\renewcommand{\tabcolsep}{12.75pt}
\begin{tabular}{llllllllllll}
\toprule	
%
\textbf{Model}
& 
Average
& Greedy
& Extrema
\\
\toprule
LSTM~\protect\cite{vinyals2015neural}
&
0.51
&0.39
&0.37
\\
\midrule
HRED~\protect\cite{serban2016building}
&
0.50
&0.38
&0.36
\\
\midrule
VHRED~\protect\cite{serban2017hierarchical}
&
0.53
&0.40
&\bf{0.38}
\\ 
\toprule
LSH-RkNN
&
{\bf 0.61$\pm$0.17}
&0.40$\pm$0.13
&0.25$\pm$0.09
\\
%
%
%
\midrule
LSH-RMM
&
\bf{0.61$\pm$0.17}
&\bf{0.41$\pm$0.13}
&0.25$\pm$0.09
\\
%
%
\midrule
LSH-RLSTM
&
0.60$\pm$0.18
&0.39$\pm$0.13
&0.24$\pm$0.09
\\
\toprule
\end{tabular}
}
%
%
%
%
\subtable[
Larry King Dataset
]
{
\centering
\renewcommand{\arraystretch}{1.4}
\renewcommand{\tabcolsep}{12.75pt}
\begin{tabular}{llllllllllll}
\toprule	
%
\textbf{Model}
& 
Average
& Greedy
& Extrema
\\
\toprule
%
%
%
%
LSTM~\protect\cite{vinyals2015neural}
&
0.71$\pm$0.24
&0.60$\pm$0.20
&0.35$\pm$0.14
\\
\midrule
%
%
%
HRED~\protect\cite{serban2016building}
&
0.71$\pm$0.25
&0.61$\pm$0.20
&0.29$\pm$0.12
\\
\midrule
VHRED~\protect\cite{serban2017hierarchical}
&
0.70$\pm$0.24
&\bf{0.72$\pm$0.25}
&\bf{0.43$\pm$0.18}
\\ 
\toprule
LSH-RkNN
&
{\bf 0.76$\pm$0.28}
&0.60$\pm$0.21
&0.34$\pm$0.15
\\
%
%
\midrule
%
LSH-RMM
&
0.73$\pm$0.28
&0.59$\pm$0.22
&0.35$\pm$0.16
\\
%
%
%
\midrule
LSH-RLSTM
&
\bf{0.76$\pm$0.27}
&0.58$\pm$0.21
&0.33$\pm$0.15
\\
\toprule
\end{tabular}
}
%
%
\caption{
Comparison between state-of-art neural network models (LSTM, HRED and VHRED) and the proposed hashing models (LSH-RkNN, LSH-RMM and LSH-RLSTM), on  three datasets -- Depression Therapy,  Twitter, and Larry King data -- using word embedding-based   similarity metrics between the actual and generated responses.       
Mean and standard deviation across samples (response pairs) are reported for all metrics, for each test set except for Twitter results with prior art  models~(LSTM, HRED, VHRED) - we used the numbers reported in \protect\cite{serban2017hierarchical}, without rerunning the models; standard deviations were not  reported   in that paper.}
\label{tab:results}
\end{table*}

\subsubsection{Mutual Information Lower Bound for Efficiency\\}
\label{sec:mi_lb}
Since computing mutual information between two high-dimensional variables can be both computationally expensive and  inaccurate if the number of samples is small~\cite{kraskov2004estimating,walters2009estimation,singh2014generalized,gao2015efficient}, we develop a (novel) lower bound on the mutual information which is easy to compute.
For derivation details,  see the supplementary material.

We will first introduce the information-theoretic quantity called \emph{Total Correlation}~\cite{watanabe1960information},  $\cT\cC(\bs{C}) = \sum\limits_j \cH(C_j) - \cH(\bs{C})$, which captures non-linear correlation among the dimensions of a random variable $\bs{C}$; given an additional  random variable $\bs{Y}$, $\cT\cC(\bs{C}: \bs{Y})$ is defined as:
 
\begin{align}
\cT\cC(\bs{C}: \bs{Y}) = \cT\cC(\bs{C}) - \cT\cC(\bs{C}|\bs{Y}).
\end{align}

\begin{theorem}[Lower Bound on Mutual Information]
Mutual information between two random hashcode variables, $\cI(\bs{C^p}: \bs{C^t})$, can be bounded from  below as follows:

{
\csizenine
\begin{align*}
\cI(\bs{C^p}: \bs{C^t})
\ge
&
\sum\limits_j{\cH(C^t_j)}
-
\cT\cC(\bs{C^t}:\bs{Y^*})
+
\sum\limits_j{\left\langle \log q(C^t_j|\bs{C^p})
\right\rangle}.
\end{align*}
}
Herein, $\cT\cC(\bs{C^t}:\bs{Y})$ describes Total Correlations within $\bs{C^t}$ that can be explained by a latent variables representation $\bs{Y}$; $q(C^t_j|\bs{C^p})$ is a proposal conditional distribution for the $j_{th}$ bit of the hashcode $\bs{C^t}$ predicted using a probabilistic classifier, like a Random Forest model.
\label{thm:mi_lb}
\end{theorem}

As discussed in \cite{greg2014discovering}, $\cT\cC(\bs{C^t}: \bs{Y}^*)$ can be computed  efficiently.
    
Note that the first two terms in the MI lower bound contribute to improving the quality of hashcodes as response representations,  maximizing entropy of each hashcode bit while discouraging redundancies between the bits, while the last term containing conditional entropies aims at improving inference of individual hashcode bits.

Moreover, one can use the proposed MI LB as an \emph{evaluation metric of the dialog quality on test data}, i.e. the alignment/relevance between the responses of two dialog agents.

\begin{table*}[tp!]
\centering
\subtable[Depression Dataset]
{
\centering
\begin{tabular}{llllllllllll}
\toprule	
&LSTM&HRED&VHRED&LSH-RkNN&LSH-RMM&LSH-RLSTM\\
%
%
%
%
%
\toprule
Appropriate~(\%)&3.7&8.6&9.5&\textbf{28.7}&24.1&25.4\\
%
%
%
%
%
\midrule
Diverse~(\%)&0.7&9.3&0.7&35.1&13.2&\textbf{41.1}\\
%
%
%
%
%
\toprule
\end{tabular}
}
%
%
\subtable[Larry King  Dataset]
{
\centering
\begin{tabular}{llllllllllll}
\toprule	
&LSTM&HRED&VHRED&LSH-RkNN&LSH-RMM&LSH-RLSTM\\
%
%
%
\toprule
Appropriate~(\%)&5.3&3.9&5.3&\textbf{31.7}&25.6&28.3\\
%
%
\midrule
Diverse~(\%)&3.3&13.3&0.0&\textbf{36.7}&10.0&\textbf{36.7}\\
\toprule
\end{tabular}
}
%
%
\caption{
Human evaluation scores on (a) Depression dataset 
(900 test samples) 
and (b) Larry King dataset~(180 test samples).
}
\label{tab:human}
\end{table*}

Also, note that mutual information criterion has been used previously to optimize kernelized hashcode representations for binary classification problems~\cite{garg2018efficient,garg2019nearly}. The approximation of mutual information proposed in those prior works are not applicable for dialog modeling.

                    
We also use a normalized metric, dividing MI LB by an upper bound on joint entropy,

\begin{align}
\cH(\bs{C^t})
&=
\sum\limits_j{\cH(C^t_j)}
-
\cT\cC(\bs{C^t})
\\
&
\leq
\sum\limits_j{\cH(C^t_j)}
-
\cT\cC(\bs{C^t}:\bs{Y^*}).
\end{align}

For $\cT\cC(\bs{C^t}|\bs{Y^*})=0$, i.e. when a latent representation $\bs{Y^*}$ is learned which explains all the Total Correlations in $\bs{C_t}$, the upper bound becomes equal to the entropy term; practically, for the case of hashcodes, learning such a representation should not be difficult, so the bound should be tight. 

%

Having learned a model to hash patient or therapist utterances, we train a Random Forest classifier, or any classifier, to infer hashcode for a therapist response given the input of a hashcode of a patient  utterance, as illustrated in \figref{fig:klsh_max_mi}. For mapping the inferred therapist hashcode to a textual response, we have two choices. One simple choice, which works well for therapeutic dialog like problems, is to select a therapist response from the training set by matching its hashcode w.r.t. the inferred hashcode. Second choice is to generate a new response given the inferred hashcode, for which we propose a novel approach in the following section.
	
\subsection{Dialogue Generation with Kernelized Hashing of N-grams}
\label{sec:gen_response}

For generation of a therapist dialogue response, given the input of its kernelized hashcode representation, we propose a new language model by generalizing the traditional approach of N-grams language modeling with kernelized hashing of N-grams.
%
%
The fundamental problem with the traditional N-grams based language models is that the approach is restricted to N-grams of short length~\cite{pauls2011faster,bengio2003neural,brown1992class}. As we increase the length of N-grams, it becomes difficult to generalize the applicability of N-grams from a training corpus to test settings due an explosion in the number of possible N-grams.
    
In light of the recent developments on learning kernelized hashcode representations of natural language~\cite{garg2018efficient,garg2019nearly}, it is possible to generalize the existing N-gram language models via kernelized hashing of N-grams. While it is low probable to exactly match an N-gram of large length w.r.t. the ones in a training corpus, it is feasible to find a match for its hashcode representation. This is because N-grams sharing relevant patterns are assigned same kernelized hashcode. Along these lines, instead of N-grams, probabilities are computed for hashcodes of N-grams per the occurrence statistics of N-grams assigned same hashcode, and so it applies for the conditional probabilities, which gives us the basic language model for generating therapist responses; see the pseudo code in \algoref{alg:response_generation}.

For hashing of N-grams, we use an unsupervised kernelized hashing model; the supervised dialog hashing model, that we proposed in the previous section, is used only for hashing of dialogue utterances, not N-grams. In \cite{garg2019nearly}, an information theoretic approach for nearly-unsupervised learning of kernelized representations is proposed for  information extraction task, that is easily extensible for unsupervised learning settings.

\section{Empirical Evaluation}
\label{sec:exp}

Several variants of the proposed   hashing based dialog model, using kNN, SVM or LSTM to build hashcodes, respectively, were evaluated    on three different datasets and compared with  three state-of-art dialog generation approaches of \cite{serban2017hierarchical,serban2016building} and \cite{vinyals2015neural}. Besides several standard evaluation metrics adopted by those approaches, we also 
report the {\em model rankings obtained by human evaluators}  via Amazon Mechanical Turk.
	
\subsection{Experimental Setup}
\subsubsection{Datasets\\}
The three datasets used in our experiments include (1) depression therapy sessions, (2) Larry King TV interviews and (3) Twitter dataset.   
    
\begin{table*}[tp!]
\centering
\centering
\begin{tabular}{llllllllllll}
\toprule	
&LSTM&HRED&VHRED&LSH-RkNN&\textbf{LSH-RkNN-Gen}\\
\toprule
%
%
Appropriate~(\%)&3.7&12.2&13.5&36.8&33.8\\
\midrule
%
%
Diverse~(\%)&0.0&4.0&4.0&80&12.0\\
\toprule
\end{tabular}
%
\caption{
Human evaluation scores on Depression dataset~(300 test samples) for our generative hashing model~(LSH-RkNN-Gen) w.r.t. the baseline models, and our response selection based hashing model~(LSH-RkNN).}
\label{tab:human_gen}
\end{table*}

The {\em depression therapy dataset}\footnote{{\small {https://alexanderstreet.com/products/counseling-and-psychotherapy-transcripts-series}}} consists of transcribed recordings of nearly 400 therapy sessions between multiple therapists and  patients. Each patient response $S_i^t$ followed by therapist response $S_i^p$ is treated as a single sample; all such pairs, from all sessions, were combined into one set of   N=42000 samples. We select 10\% of the data randomly as a test set (4200 samples), and then perform another random 90/10 split of the remaining  38,000 samples  into  training and validation subsets, respectively. We follow the experimental setup from prior work cited above when comparing the respective neural network models with our hashing based approaches: all models    are  trained  only once using the same training and validation datasets, and evaluated on the same test set. However, for our hashing model metrics introduced below,   we average the estimates over  10 random subsets using  95\% of test samples each time.
    
The {\em Larry King dataset}~\footnote{  {\small http://transcripts.cnn.com/TRANSCRIPTS/lkl.html}}
contains transcripts of interviews with the guests of  TV talk shows, conducted by   Larry King, the host. Similarly to the depression therapy dataset, we put together all pairs of guest/host responses from 69 sessions into a single set of size 8200. The data are split into training, validation and test subsets as described earlier.

Next, we experimented with the   {\em Twitter Dialogue Corpus} \cite{ritter2010unsupervised}.
Considering the original tweet and the following comments on it, in the same session, the task is to infer the next tweet.  Note that we consider all utterances preceding that tweet as one long utterance, i.e. as the first  ``response'' $S_i^A$, mapped to one hashcode,   while the next tweet is the second ``response''  $S_i^B$, which is different from the approach of \cite{serban2017hierarchical} we compare with, where   the previous   utterances in a session are explicitly viewed as a sequence.
The number of tweet sessions (each viewed as a separate sample, i.e.  $\{ S_i^A, S_i^B \}$ pair of responses),   in training, validation, and test subsets are, respectively, 749060, 93633, and 93633.

\subsubsection{Task\\} 
For all datasets, the task is to train a model on a set of training samples, i.e. response pairs $(S_i^A,S_i^B)$, where $S_i^A$ is a response of person A, followed by the corresponding response of  a person B. Then each test sample is given as a response of person A, and the task is to generate the response    of a person B.

\subsubsection{Hashing Models\\}

{\em Step 1: Representation Learning. }	
We evaluate three different hashing models: the first two, based on    kernel locality sensitive hashing (KLSH) \cite{joly2011random,garg2018efficient}, are  called \emph{LSH-RMM} and \emph{LSH-RkNN}, and  use, respectively, Max-Margin (SVM)  classifier  (with C=1 parameter) or kNN classifier~(k=1), to compute each   hash function.
The third model, \emph{LSH-RLSTM}, uses LSTM  for hash function computation.
We use hashcode vectors of dimensionality H=100. 
For \emph{LSH-RkNN} and \emph{LSH-RMM}, we use as a reference set a  random subset of M=300 samples from the training dataset, to reduce the computational complexity of training those models, but for \emph{LSH-RLSTM}  we use the whole training dataset as a reference set.
Parameters $\bs{\theta}$ for  LSH models are obtained by   maximizing the proposed MI LB criterion.
%

{\em Step 2: Hashcode prediction.}
We now map  all responses, of both participants A and B,  in both training and test sets, to the corresponding hashcodes using one of the above hashcode-based representation models.  
Next, to predict the   response hashcode of a person B given a hashcode of a person A, we train  separate Random Forest (RF) classifiers (each containing 100 decision trees) for each  hashcode bit (i.e. 100 such RF classifiers, since H=100).
 
 {\em Step 3: Textual response generation.}
 Given a hashcode of a response inferred by RF classifier above, mapping it to an actual text can be performed in multiple ways; by default, we find a match of the generated hashcode in the set of all hashcodes corresponding to the person B responses in our training data as this approach is highly suitable for depression therapy like dialogues. In addition, we also provide human evaluation for generating new responses using the N-grams hashing based language model proposed in \secref{sec:gen_response}.
%

\subsubsection{Baseline: Neural Network Dialog Generation Models\\}
We compare our dialog generation method with the state-of-art {\em VHRED} approach of \cite{serban2017hierarchical}, as well as with the two other approaches, {\em HRED} \cite{serban2016building}, and {\em LSTM} \cite{vinyals2015neural}, also used as  baselines in the VHRED paper. We adopt  the same hyperparameter settings as those used in   \cite{serban2017hierarchical}. For the  Twitter dataset, we compare with  the results presented in the above paper, while on the other two datasets, we train the above models ourselves. The vocabulary size for the input is set via grid search between values 1000 to 100000. The neural network structures are chosen by an informal search over a set of architectures and we set maximum gradient steps to 80, validation frequency to 500 and step-size decay for SGD is 1e-4.

\subsubsection{Evaluation Metrics\\}

{\em Embedding-based metrics.}
We compare our methods with the state-of-art neural network approaches listed above using three word embedding-based topic similarity metrics -  {\em embedding  average},    {\em embedding greedy}, and {\em embedding  extrema} \cite{liu2016not}, adopted  by \cite{serban2017hierarchical}. Following the prior art, we used  Google News Corpus to train the embeddings.
%
The mean and standard deviation statistics for each metric are computed over 10 runs of the experiment, as mentioned above.




\begin{table*}[tp!]
\centering
%
\subtable[
Depression Therapy Dataset
]
{
\centering
\begin{tabular}{llllllllllll}
\toprule	
\textbf{Model}
& MI LB~(Shuffled)
& NMI LB
&HIA~(Baseline)
\\
\toprule
LSH-RkNN
%
&$12.8\pm$0.5~(6.1$\pm$0.2)
&0.57~(0.27)
&0.87$\pm$0.16~(0.82$\pm$0.18)
\\
%
%
%
%
%
%
\midrule
LSH-RMM
%
%
&{\bf13.7$\pm$0.2~(-1.0$\pm$0.3)}
&{\bf0.39~(0.0)}
&{\bf0.68$\pm$0.10~(0.59$\pm$0.10)}
\\
%
%
%
%
%
%
\midrule
LSH-RLSTM
&20.3$\pm$0.3~(10.6$\pm$0.2)
&0.76~(0.40)
&0.82$\pm$0.19~(0.79$\pm$0.21)
\\
\toprule
\end{tabular}
}
\subtable[
Twitter Dataset
]
{
\centering
\begin{tabular}{llllllllllll}
\toprule	
%
%
%
\textbf{Model}
&
MI LB~(Shuffled)
& NMI LB
&HIA~(Baseline)
\\
\toprule
LSH-RkNN&
%
17.3$\pm$0.1~(9.1$\pm$0.1)
&0.75~(0.40)
&0.85$\pm$0.18~(0.81$\pm$0.21)
\\
%
%
\midrule
LSH-RMM&
{23.4$\pm$0.2~(11.2$\pm$0.3)}
&{\bf0.71~(0.34)}
&{\bf0.80$\pm$0.14~(0.72$\pm$0.16)}
\\
%
%
\midrule
LSH-RLSTM&
{\bf 41.9$\pm$0.1~(24.3$\pm$0.1)}
&0.83~(0.48)
&0.69$\pm$0.16~(0.63$\pm$0.16)
\\
\toprule
\end{tabular}
}
\subtable[
Larry King Dataset
]
{
\centering
\begin{tabular}{llllllllllll}
\toprule	
%
\textbf{Model}
& 
MI LB~(Shuffled)
& NMI LB
& HIA~(Baseline)
\\
\toprule
LSH-RkNN&
9.4$\pm$0.3~(5.1$\pm$0.3)
&0.63~(0.34)
&0.91$\pm$0.14~(0.89$\pm$0.16)
\\
%
%
%
\midrule
LSH-RMM&
%
%
22.4$\pm$0.9~(4.5$\pm$1.2)
&{\bf0.62~(0.13)}
&{\bf0.69$\pm$0.11~(0.59$\pm$0.10)}
\\
%
%
%
\midrule
LSH-RLSTM
&{\bf 48.9$\pm$0.3~(28.4$\pm$0.5)}
&0.80~(0.47)
&0.62$\pm$0.10~(0.54$\pm$0.07)
\\
\toprule
\end{tabular}
}
%
%
\caption{
Hashcode models quality as measured by the alignment between the hashcodes of person A and person B responses  (mutual information lower bound and its normalized version), as well as by the predictability of B's responses given A's response. 
%
%
%
%
}
\label{tab:results2}
\end{table*}

\noindent{\em Human evaluation.}
Using Amazon Mechanical Turk, we obtained model rankings from 108 human readers (annotators).  For each test sample, we showed to the reader all   responses produced the six models evaluated here, in random order for each instance, and without specifying which model produced which response. We asked the annotator to choose two most appropriate responses; then,  for each model,  we computed across all test samples the  percent of the readers who voted for that model.  Furthermore, in a separate session, we also asked the person which model produced the most diverse responses; in this case we listed the model's names and associated responses.

\noindent{\em Information-theoretic metrics.}
The second group of metrics directly evaluates the quality of hashcodes obtained using our models.  For each method, we will report the  proposed MI lower bound~(\emph{MI LB}) from \thmref{thm:mi_lb}, as well as its normalized version~(NMI LB). For each of the metrics, \emph{higher values} mean better performance. 
We also report the {\em hashcode inference accuracy (HIA)},  i.e. the accuracy of predicting hashcode bits of response B using  RF classifiers. We also obtain the baseline accuracy~(\emph{Baseline}), using a trivial classifier that always chooses the most-frequent class label.

%

\subsection{Results}
\subsubsection{Computational Efficiency\\}
First of all,  we observed that {\em the hashing models are much more computationally efficient than the neural network approaches we compared with}:   it takes from several days (for smaller datasets such as Depression therapy and Larry King dataset containing about 42,000 and 8,200 samples) to more than two weeks (on Twitter, with roughly 749,000 samples) to train a neural network model, even on a 1000-core GPU, whereas an LSH model is typically optimized within several hours (e.g., less than 5-6 hours for Twitter),  on a 16-core CPU.

\subsubsection{Embedding-based Metrics\\}
Table \ref{tab:results} summarizes performance of all methods with respect to the embedding-based metrics, with the best results for each metric/column shown in boldface. Overall, the proposed hashing approaches are quite  competitive with the neural net  methods in terms of those metrics as well: on all three datasets, our LSH methods always outperform their competitors in terms of the {\em average} similarity metric (which is the most intuitive among the three metrics in terms of reflecting the topic similarity between the true and system-generated responses); moreover, our methods also achieve best performance w.r.t. the {\em greedy} metric on Twitter dataset, and w.r.t. {\em extrema} metric on the Depression dataset.  
    
\subsubsection{Human Evaluation\\}
We  also performed an extensive human evaluation of the dialog generation quality (as described earlier in section 5.1), on two relatively smaller datasets, Depression and Larry King. The results are shown in Table \ref{tab:human}a and \ref{tab:human}b, respectively.  {\em The hashcode-based approaches considerably outperform the neural net models, by one-two orders of magnitude,  in terms of the responses being both more appropriate and more diverse}\footnote{See several examples in the supplementary material.}. Amongst the three hashing models, LSH-RkNN seem to be performing the best.
		
Further, we performed another human evaluation of LSH-RkNN, for a comparison between response generation vs response selection, and their comparison w.r.t. the baseline neural models. As we observe in \tabref{tab:human_gen}, both response selection and response generation seem to give equally good responses in terms of appropriateness, and significantly outperform the neural baselines. Low diversity of generated responses remains an issue, as also known to be the case for neural models.

\subsubsection{Information-theoretic Metrics\\}
Next, we took a deep dive into evaluation of hashing approaches with respect to how well they actually model the alignment between the responses; the results are summarized in Table \ref{tab:results2}, presenting the mutual information lower bound (MI LB), normalized MI LB, and hashcode inference accuracy (HIA) as discussed before; note that besides presenting chance-level baseline classifier accuracy (i.e., selecting the most-frequent class), we also present a similar baseline for MI LB and NMI LB metrics, using  randomly permuted (shuffled) pairings between responses of person A and person B. We now look at the  difference between each  metric and its corresponding baseline, shown in parenthesis, and highlight the largest, most significant differences between the means of both metrics in boldface. 
    
Overall, {\em LSH-RMM} method appear to be the best on all datesets  in terms of most significant improvement over the baseline in terms of normalized MI LB and prediction accuracy, although {\em LSH-RLSTM} is the best in term of the raw MI LB on Twitter and Larry King datasets.

\section{Related Work}
\label{sec:rw}
%
Therapy chatbots, such as  Woebot \cite{fitzpatrick2017delivering}  and similar systems,  
are becoming increasingly popular; however, these agents have  limited ability to  understand free text and have to resort to a fixed set of preprogrammed responses to choose  from~\cite{di2017chatbots,ly2017fully,schroeder2018pocket,morris2018towards,hamamura2018standalone}. (Also, see \cite{jurafsky2014dialog} for an overview.)
    
For dialogue modeling in general domains, several recently proposed neural network based approaches are considered state-of-art ~\cite{serban2016generative,serban2017hierarchical,shao2017generating,asghar2017deep,wu2017sequential,li2017adversarial,zhao2018unsupervised,wu2018learning,xu2018diversity,park2018hierarchical,jiang2018sequence,tao2018get,xing2018hierarchical,chen2018hierarchical,shi2018toward,du2018variational}. However, those approaches usually require very large training datasets, unavailable in many practical applications; 
furthermore, they are not typically explored in  dialogue settings such as therapy including very long responses (up to tens of thousands of words).  Also, evaluating the effectiveness of the therapist's response  requires some  notion of relevance (e.g., mutual information) which goes beyond the standard measures of its semantic features~\cite{papineni2002bleu,liu2016not,li2017neural,lowe2017towards,li2017adversarial}.

Unlike  task-driven dialogue~\cite{zhai2014discovering,wen2016network,althoff2016large,lewis2017deal,he2018decoupling},   an immediate response quality metric may not   be available in our settings, since the effect of therapy is harder to evaluate and multiple sessions are often required to achieve the desired outcome.
Attention to specific parts of the response, as well as background knowledge, explored in neural network based dialogue modeling~\cite{kosovandialogue} can be helpful in therapeutic dialogues; those aspects  are, to some extent, implicitly captured by learning the hashing models. 
%
%
    
In \cite{he2018decoupling}, an approach is presented to a task-driven (e.g., negotiation) dialogue which consists in  mapping a response to an ordered list of rules, where each rule represents a task-specific intent; this does not apply to our more open-ended dialogues without the specific tasks.
    
While mutual information has been previously considered in dialogue modeling during testing \cite{li2016diversity}, it was not used as a model selection criterion for  learning representations.
%
    
Finally,  there are multiple approaches for  estimating mutual information   from data
\cite{kraskov2004estimating,koeman2014mutual,singh2014generalized,gao2015efficient}. However, these estimators are highly expensive  in high-dimensional settings, and can be quite inaccurate when the number of samples is  small. There is a recent approach of neural estimation of mutual information on two high dimensional continuous variables, though not applicable to discrete variables like hashcodes~\cite{belghazi2018mine}. For discrete variables, theoretical analysis has been limited to one dimensional case~\cite{jiao2017maximum}. Previously, several mutual information lower bounds have been proposed for classification problems~\cite{chalk2016relevant,gao2016variational,alemi2017deep}, assuming one-dimensional class label; unfortunately, they do not apply in our setting where the predicted response is a high-dimensional  vector. 

\shrink{ 
\todo{for Sahil, add references suggested by AISTATS reviewers}

\todo{For Sahil, mention recent estimators for mutual information, such as MINE, which are relevant for continous variables whereas we have binary~(hashcode) representations in this work.}
\todo{For Sahil, add recent citaions from 2018, replacing older ones.}
} 

%
%
%
%
%
%
%
%
%
%
%
%

\section{Conclusions}
%
This paper introduces a novel approach to dialogue modeling where responses of both participants are represented by kernelized hashcodes. Furthermore, a novel lower bound on Mutual Information is derived and used as a hascode-based model-selection criterion in order to facilitate a better alignment in collaborative dialogue, as well as predictability of responses. Our empirical results consistently demonstrate superior performance of the proposed approach over state-of-art neural network dialogue models in terms of both computational efficiency and response quality. For next steps, we plan to further improve the approach by choosing a better response from a larger corpus, implement more sophisticated hashcode-to-text generative models, and extend the model to the dynamics of a dialogue beyond immediate responses. Additionally, we will exploit the vast literature on operationalized assessment of several psychological qualities such as the Beck Depression~\cite{BDI} and Beck Anxiety~\cite{BAI} Inventories, as well as features of dialogue such as the Working Alliance Inventory~\cite{WAI}, in order to derive better hashcode representations and predictive models. In this sense, we expect that the lessons learned in richly evaluated therapy sessions can be extrapolated, mutatis mutandis, beyond the realm of mental health.     

{
\bibliography{references}
\bibliographystyle{aaai}
}

\end{document}